\documentclass{ieeeaccess}
\usepackage{cite}
\usepackage{amsmath,amssymb,amsfonts}
\usepackage{algorithmic}
\usepackage{graphicx}

\usepackage{gensymb}
\DeclareGraphicsExtensions{.pdf,.jpeg,.png}
\usepackage[autostyle]{csquotes}

\usepackage{bbm} 
\usepackage{booktabs}
\usepackage{multirow,bigdelim}

\def\BibTeX{{\rm B\kern-.05em{\sc i\kern-.025em b}\kern-.08em
    T\kern-.1667em\lower.7ex\hbox{E}\kern-.125emX}}
\begin{document}
\history{Date of publication xxxx 00, 0000, date of current version xxxx 00, 0000.}
\doi{10.1109/ACCESS.2017.DOI}

\title{Selective Distillation of Weakly Annotated GTD for Vision-based Slab Identification System}
\author{
	\uppercase{Sang Jun Lee}\authorrefmark{1}, \IEEEmembership{Member, IEEE},
	\uppercase{Sang Woo Kim}\authorrefmark{1}, \IEEEmembership{Member, IEEE},
	\uppercase{Wookyong Kwon}\authorrefmark{2},
	\uppercase{Gyogwon Koo}\authorrefmark{1}, \IEEEmembership{Member, IEEE},	
	\uppercase{Jong Pil Yun}\authorrefmark{3}, 
	}
\address[1]{Department of Electrical Engineering, POSTECH, Pohang 37673, Republic of Korea (e-mail: lsj4u0208@postech.ac.kr)}
\address[2]{POSCO (Pohang Iron and Steel Company), Pohang 37859, Republic of Korea}
\address[3]{Aircraft System Technology Group at Korea Institute of Industrial Technology (KITECH), Daegu 42994, Republic of Korea}

\markboth
{Author \headeretal: Preparation of Papers for IEEE TRANSACTIONS and JOURNALS}
{Author \headeretal: Preparation of Papers for IEEE TRANSACTIONS and JOURNALS}

\corresp{Corresponding author: Jong Pil Yun (e-mail: rebirth@kitech.re.kr)}

\begin{abstract}
This paper proposes an algorithm for recognizing slab identification numbers in factory scenes.
In the development of a deep-learning based system, manual labeling to make ground truth data (GTD) is an important but expensive task.
Furthermore, the quality of GTD is closely related to the performance of a supervised learning algorithm.
To reduce manual work in the labeling process, we generated weakly annotated GTD by marking only character centroids.
Whereas bounding-boxes for characters require at least a drag-and-drop operation or two clicks to annotate a character location, the weakly annotated GTD requires a single click to record a character location.
The main contribution of this paper is on selective distillation to improve the quality of the weakly annotated GTD.
Because manual GTD are usually generated by many people, it may contain personal bias or human error.
To address this problem, the information in manual GTD is integrated and refined by selective distillation.
In the process of selective distillation, a fully convolutional network is trained using the weakly annotated GTD, and its prediction maps are selectively used to revise locations and boundaries of semantic regions of characters in the initial GTD.
The modified GTD are used in the main training stage, and a post-processing is conducted to retrieve text information.
Experiments were thoroughly conducted on actual industry data collected at a steelmaking factory to demonstrate the effectiveness of the proposed method.
\end{abstract}


\titlepgskip=-15pt

\maketitle

\section{Introduction}
Computer vision techniques have been used in various industry fields including driver assistant systems \cite{shen2018nighttime, prioletti2013part, o2010rear, yuan2018robust}, unmanned aerial vehicles \cite{garcia2017recognition, al2017survey} and medical image analysis \cite{sompong2017efficient, nabizadeh2017automatic}.
Recently, convolutional neural network (CNN) has received big attention as the most promising method for these computer vision tasks after the breakthrough in image classification \cite{krizhevsky2012imagenet} and object detection \cite{szegedy2013deep}.
In the area of computer science, many researches were conducted on network architectures \cite{szegedy2015going, he2016deep}, regularization methods \cite{srivastava2014dropout, ioffe2015batch} and effective cost function \cite{niu2016ordinal} to improve the performance of deep learning algorithms on well-defined problems.
However, solving a real-world problem arises another kind of concerns such as data acquisition framework and problem formulation for satisfying system requirements.
For example, a deep-learning method for fire surveillance system which covers a variety of fire accidents was proposed in \cite{muhammad2018efficient}, and automatic inspection system for nuclear power plant components was established by using a small number of crack data \cite{chen2018nb}.
To predict remaining useful life of bearings, data acquisition procedure and training method of a deep-learning algorithm was proposed in \cite{ren2018prediction}.
For autonomous driving systems, challenging corner cases such as detection of pedestrians and turn signals at night time were considered to prevent accidents \cite{kim2018pedestrian, chen2017turn}.

This paper is about weakly annotated ground truth data (GTD) and its training method to establish a deep-learning based system for solving a real world problem.
To train a supervised learning algorithm, GTD are manually generated in the process of data preparation, and the quality GTD is closely related to the performance of the algorithm.
Although GTD are conventionally regarded as \emph{true}, labeling criterion may be inconsistent because this manual work is usually conducted by many people involving individual interpretations.
For example, categories of ambiguous objects in images are determined subjectively, and object locations are annotated by loose or tight bounding-boxes depending on personal bias.
Furthermore, annotating large-scale data requires laborious manual work.

There have been several attempts to reduce manual work in labeling process.
For scene segmentation, vertices of polygons were annotated to represent detailed shapes of objects, and these annotations were converted to pixel-level class labels \cite{russell2008labelme, lin2014microsoft}.
A semi-automatic annotating method was proposed in \cite{castrejon2017annotating} to generate vertices of polygons based on object bounding-boxes.
To reduce manual work for generating GTD, weakly annotating methods that rely on scribbles \cite{lin2016scribblesup} or points \cite{bearman2016s} were proposed.
As training methods, weakly supervised learning algorithms were proposed in \cite{oquab2015object, zhou2016learning} to conduct object localization by using image-level category labels.
Inspired by these previous researches, in this paper, we propose a weakly annotating method for scene text recognition and a subsequent distillation algorithm to refine the initial GTD.

The objective of this paper is to develop an algorithm for recognizing handwritten slab identification numbers (SINs) in factory scenes.
In steel industry, slabs have similar shapes in spite of different compositions of alloying elements.
Therefore, a SIN is inscribed on each slab to monitor individual products, and automatic recognition of SINs is an important prerequisite for the automation of steel-making processes.
Recognizing variable number of SINs with near-perfect performance is a challenging problem.
Background of factory scenes is complicated, and characters are usually deformed or blurred during high-temperature manufacturing processes.
To address these problems, a computer vision system to obtain factory scenes with slab information was proposed in \cite{choi2012localizing}, and algorithms for recognizing machine-printed SINs were presented in \cite{lee2017localization} and \cite{lee2017end}.
However, SINs cannot be automatically inscribed in some situations (e.g. absence or malfunction of a marking machine), and several SINs are inscribed by manually.
These manually inscribed SINs are called \emph{handwritten SINs}.
The recognition of handwritten SINs is a more difficult problem compared to the recognition of machine-printed SINs due to various types and sizes of handwritten characters.
Fig.~\ref{fig:example_scenes} shows example factory scenes with handwritten SINs.


\Figure[t!](topskip=0pt, botskip=0pt, midskip=0pt)[width=\linewidth]{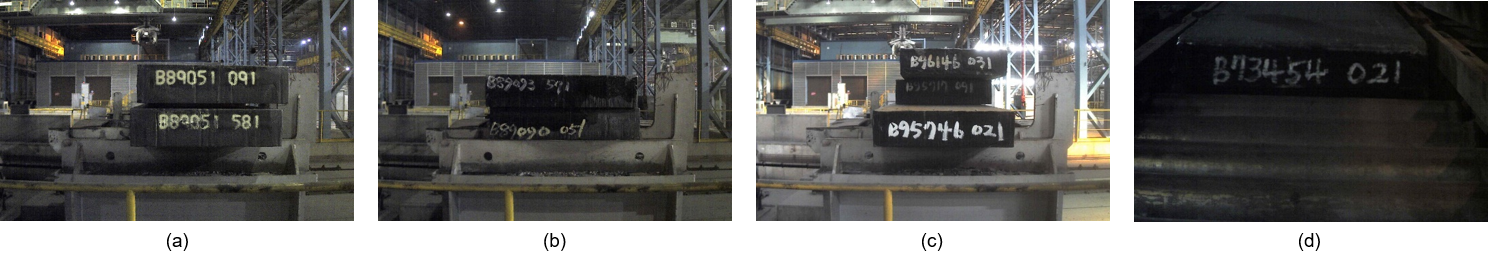}
{Example factory scenes with SINs. (a) contains machine-printed SINs, and (b)-(d) contain handwritten SINs.\label{fig:example_scenes}}

In this paper, fully convolutional network (FCN) is used to recognize handwritten SINs in factory scenes.
FCN is a deep learning model that is composed of convolution layers without any fully-connected layer, and it can be trained by using pixel-level semantic meaning of input images.
Most GTD for scene text recognition are based on bounding-boxes which require a drag-and-drop operation or at least two clicks to represent a character location.
To alleviate manual work in labeling process, we annotated data with weak supervision by marking only character centroids, and this method requires a single click for annotating a character location.
Semantic region for a character was defined as an isolated circular region centered at each annotated point, and categorical label was assigned in the semantic region to generate initial GTD.

The main contribution of this paper is on selective distillation of the initial GTD.
In labeling process, people have varying level of precision or criterion for annotating character centroids.
To obtain consistent GTD, the information of entire training data was integrated into a FCN by training with the weakly annotated GTD.
By applying a post-processing to verify predictions of the initially trained network, distillation was selectively conducted to revise locations and boundaries of the initially annotated character regions; we call this process \emph{Selective Distillation}.
These modified GTD were used in the main training process to build a recognition algorithm for handwritten SINs in factory scenes.
Experimental results demonstrate that the process of selective distillation is effective to improve recognition performance with a large margin.
The remaining sections are organized as follows.
Section 2 presents related work and Section 3 explains the proposed method including the process to make weakly annotated GTD, recognition algorithm, and selective distillation for revising the initial GTD.
Section 3 presents experiment results, and Section 4 contains conclusions.

\section{Related work}
\subsection{Knowledge distillation}
Distillation for a deep neural network was firstly proposed in \cite{hinton2015distilling}.
In \cite{hinton2015distilling}, a shallow model (student network) was trained to mimic responses of a deeper model (teacher network).
Different to one-hot encoded labels, responses of a teacher network contain probabilistic information of both true and false categories, and this additional information about false categories helps a student network to achieve close performance to its teacher network despite the low complexity of the student network.
The main feature of knowledge distillation is using responses of a teacher network to train a student network, and this approach was used for network minimization \cite{yim2017gift}, domain adaptation \cite{gupta2016cross} and semi-supervised learning \cite{tarvainen2017mean}.
Recently, distillation was employed to object detection \cite{chen2017learning} and defense mechanism for adversarial samples \cite{papernot2016distillation}.
Whereas most previous works focused on classification problems, this paper extends the concept of distillation for addressing a problem of scene text recognition, and it was applied to a real industry problem.

\subsection{Scene text recognition}
Recognition of handwritten SINs belongs to the problem of scene text recognition, and most of recent algorithms employed deep neural networks.
For text localization, region proposals such as maximally stable extremal regions were used as initial cues \cite{wang2018crf, ma2017arbitrary, gomez2017textproposals}, and contexts of CNN features were integrated by using recurrent neural networks (RNNs) to refine word locations \cite{tang2017scene, tian2016detecting, wu2017deep}.
Many recent algorithms are based on FCNs to localize text information \cite{shi2017detecting, zhang2016character, zhang2016multi, vo2016dense}.
For recognizing word images, unified architectures of CNN and RNN were proposed in \cite{shi2017end} and \cite{he2016reading}.
As another approach, end-to-end recognition algorithms were proposed to jointly optimize localization and recognition processes \cite{wang2012end, jaderberg2016reading, su2017accurate, buvsta2017deep}.
Although these previous methods worked well on several benchmark datasets, word-level recognition algorithms are not applicable to recognize SINs because products have different serial numbers.

The most related algorithm to our work was proposed by \cite{lee2018recognition}, and it used an approach of semantic segmentation \cite{shelhamer2017fully} for recognizing machine-printed SINs in factory scenes.
In \cite{lee2018recognition}, semantic region of each character was defined as a rectangular region in ground truth, and individual characters were recognized by using characteristics of machine-printed characters such as consistent aspect ratios and similar distances between characters.
However, this algorithm is not applicable to recognize handwritten SINs due to irregular spacing between handwritten characters and various aspect ratios.

\section{Proposed Method}

\subsection{Weakly annotated GTD}
For scene text recognition, character locations are usually represented as bounding-boxes.
A bounding-box is characterized by a 4-tuple of (x-position, y-position, width, height), and it requires at least a drag-and-drop operation or two clicks for indicating a character location.
The proposed method to make weakly annotated GTD has a merit that it requires a single click to indicate a character position.

The procedure for generating weakly annotated GTD is presented in Fig.~\ref{fig:GTD_procedure}.
In labeling process, character centers were recorded by 2-tuples of (x-position, y-position) with their corresponding character classes.
These records were converted to 2-dimensional data by defining semantic regions of characters as shown in Fig.~\ref{fig:GTD_procedure}(b).
Because an image contains very few pixels that correspond to character centers, assigning background category for the remaining pixels causes highly imbalanced data.
Therefore, a semantic region for a character was defined as a circular region centered at its manually annotated point, and the radius of the \emph{i-th} character region was decided by
\begin{equation}
r_i = (\min_j d_{ij})/4.
\end{equation}
In (1), $d_{ij}$ is the distance between centers of the \emph{i-th} and \emph{j-th} characters, and $\min_j d_{ij}$ is the minimum distance to the other centers from the \emph{i-th} character center.
By dividing $\min_j d_{ij}$ by 4, it is guaranteed that character regions are separated to each other.
If characters in a SIN are similar in size, the diameter of a circle is close to a distance between adjacent character regions.
Text information represented by isolated regions can be retrieved by using a simple post-processing such as k-means algorithm.

In weakly annotated GTD, whose width and height are same as factory scenes, each pixel value indicates a category of the identical point in the corresponding scene as presented in Fig.~\ref{fig:GTD_procedure}(c).
Class information for each color in a ground truth image is depicted in Fig.~\ref{fig:GTD_procedure}(d), and, for example, background category was labeled by $0$.
In our industry problem, the first character of a SIN represents a production line in an actual steelworks, and the remaining characters are numerals that characterize a product.
Because the dataset was collected at a production line which was represented by `B', the number of categories is 12 including the background class.
These weakly annotated GTD are used as initial GTD to train a FCN, and selective distillation was subsequently conducted to revise the initial GTD.


\Figure[t!](topskip=0pt, botskip=0pt, midskip=0pt)[width=\linewidth]{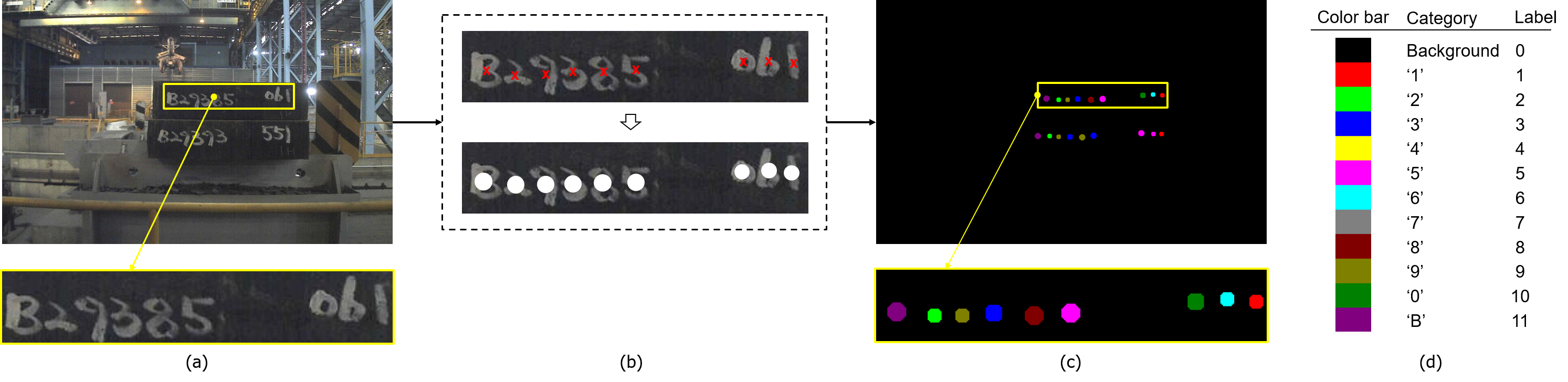}
{Process for making weakly annotated GTD. (a) An example factory scene. (b) Manually annotated character centers (red marks) and their semantic regions (white blobs). (c) Weakly annotated GTD based on semantic regions of characters. (d) Class information for each color in (c).\label{fig:GTD_procedure}}

\subsection{FCN Architecture}

Development of an appropriate deep learning model is an important task to solve a real industry problem.
Although deep neural networks using big data usually show better performance than shallow networks, an excessively complex model can cause overfitting problem if training data is not sufficient.
This section presents FCN architectures that were used to train a mapping from input factory scenes to the corresponding ground-truth images.
In this paper, several FCNs were developed for the recognition of SINs, and their performance is analyzed to obtain a base FCN model.
The base FCN model and two auxiliary models with different complexities are presented in Fig.~\ref{fig:FCN_models}.
These two auxiliary models were named as \emph{Shallow model} and \emph{Deep model}, and the performance and complexity of these models are analyzed in Section~\ref{sec:analysis_FCNs}.
Although this paper employed VGG19 \cite{simonyan2014very} to compute intermediate feature maps, deeper networks such as ResNet \cite{he2016deep} or more recent architecture for scene segmentation such as U-net \cite{ronneberger2015u} and deeplab \cite{chen2018deeplab} can be utilized to construct the base network, and selective distillation can be applied to improve the performance in the same way.

The base network was constructed in fully convolutional manner for pixel-level classification.
Deconvolution layers were used for parametric upsampling of feature maps, and these upsampled feature maps were combined to make an output prediction map.
Fig.~\ref{fig:FCN_models}(a) presents detailed components of the base FCN model.
Convolution layers and pooling layers are depicted by blue and green rectangles, and two or four convolution layers are tied into a convolution block (gray dashed rectangles).
Kernels with the size of $3 \times 3$ were used in convolution layers, and a rectified linear unit (ReLU) was followed for each convolution layer.
Pooling layers conducted max pooling operations on $2 \times 2$ windows.
Feature maps of the last convolution layer in each convolution block were upsampled by deconvolution layers (red rectangles).
The number of kernels for convolution layers is presented on each convolution block, and upscaling ratios of deconvolution layers are presented below red rectangles.
Kernel sizes of the deconvolution layers were two times larger than their upscaling ratios.
The output of the last deconvolution layer is denoted by $F$, and its depth is same as the number of categories including background class.
The output prediction map of the FCN can be obtained by applying an argmax operation for $F$ in depth direction, and it is denoted by
\begin{align}
O=\{ o_{ij}: ~i=1, \cdots, H, ~j=1, \cdots, W \}.
\end{align}
In (2), $H$ and $W$ are height and width of the input image, and $o_{ij} \in \{0, \cdots, K\}$ is the predicted category at the corresponding point $x_{ij}$ in the input image, where $K$ is the number of character classes.
As presented in Fig.~\ref{fig:GTD_procedure}(d), $o_{ij}=0$ implies that $x_{ij}$ is classified into the background category, and a nonzero value of $o_{ij}$ means that the predicted category of $x_{ij}$ is a character.
In the training process, cross-entropy loss between soft-max version of $F$ and one-hot encoded version of GTD was minimized by using Adam optimization \cite{kingma2014adam}.


\Figure[t!](topskip=0pt, botskip=0pt, midskip=0pt)[width=3.3in]{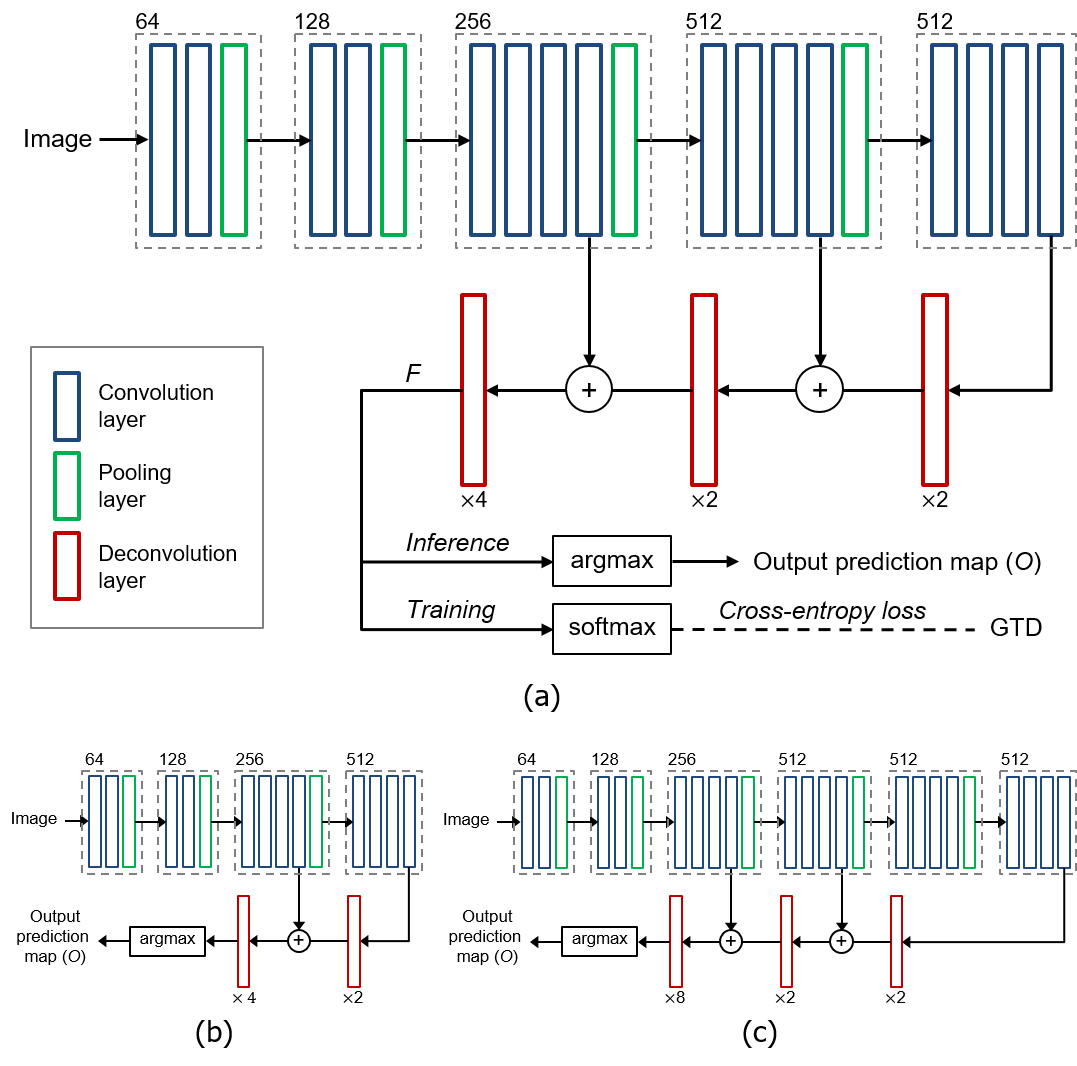}
{The base FCN model and two auxiliary architectures. (a) Base FCN model. (b) Shallow model. (c) Deep model.\label{fig:FCN_models}}

\subsection{Post-processing}

A post-processing was applied for output prediction maps to retrieve text information or to conduct selective distillation.
The post-processing is composed of two steps; the number of slabs and their vertical locations are estimated in the first step, and each SIN is recognized in the second step.
Because cross-sections of slabs are rectangular, characters in a SIN are roughly placed in the horizontal direction, and this feature was used to estimate vertical locations of slabs.
At first, the number of nonzero elements in the \emph{i-th} row of $O$ was computed by
\begin{equation}
N_i = \Sigma_{j=1}^W \mathbbm{1}_{o_{ij}>0}{(o_{ij})},
\end{equation}
where $\mathbbm{1}_A{(x)}$ is an indicator function such that $\mathbbm{1}_A{(x)}=1$ for $x \in A$.
Because $o_{ij}$ is a pixel-level prediction which has same categorical meaning to the label information depicted in Fig.~\ref{fig:FCN_models}(d), a nonzero element of $O$ implies that the identical point in the corresponding scene is in a character region.
Therefore, $N_i$ is the number of points that were classified into a character category in the \emph{i-th} row of the input image.
The set of vertical indexes of character candidates was obtained by 
\begin{equation}
I = \{i: N_i>0 \}.
\end{equation}
Let $I^{(l)}$ be the \emph{l-th} partition with connected indexes of $I$, then $I^{(l)}$ represents the set of vertical indexes of the \emph{l-th} SIN candidates.
For recognizing the \emph{l-th} SIN, the sub-region corresponding to the vertical indexes in $I^{(l)}$ was extracted from the output prediction map $O$.

The second step is to recognize individual characters in each sub-region of $O$.
Let $O^{(l)}$ be the \emph{l-th} sub-region of $O$ such that 
\begin{equation}
O^{(l)} = \{o_{ij}: ~i\in I^{(l)}, ~j = 1, \cdots, W\}.
\end{equation}
Because a SIN is compose of nine characters, k-means algorithm with the $k$ of 9 was conducted to cluster nonzero responses in $O^{(l)}$.
The category with the maximum frequency in each cluster was transcribed to retrieve character information.
If there exist even one missing character in a SIN, identification of the slab is impossible.
Therefore, SIN candidates which contain less than nine nonzero elements were filtered out to reduce false predictions.
The overall procedure for recognizing SINs in a factory scene is described in Fig.~\ref{fig:post-processing}.

\Figure[t!](topskip=0pt, botskip=0pt, midskip=0pt)[width=\linewidth]{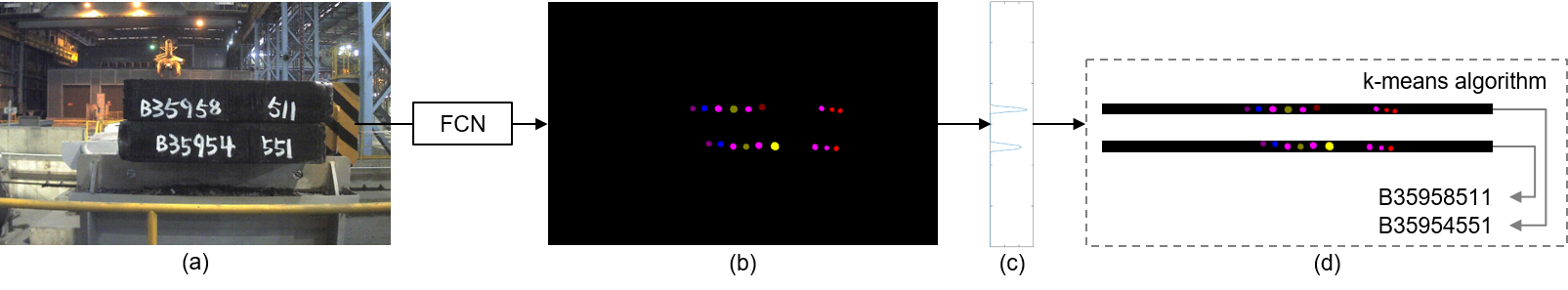}
{The procedure for recognizing SINs. (a) Example factory scene. (b) Output prediction map of the FCN. (c) The number of non-zero responses in each row of the output prediction map. (d) Sub-regions of the output prediction map and their recognition results by using k-means algorithm.\label{fig:post-processing}}

\subsection{Selective Distillation}
During the generation of weakly annotated GTD, people had varying level of precision due to human error.
Furthermore, annotators had inconsistent criteria for annotating character centroids in ambiguous images, and personal bias or interpretation was involved in labeling process.
To address this problem, distillation was employed to revise locations and boundaries of semantic regions for individual characters.
Although GTD were labelled involving interpretation of one person, a prediction of a network was related to interpretations of the entire annotators.
We believed that this overall interpretation was more effective than each individual interpretation to train a network.

To modify the weakly annotated GTD, prediction maps of an initially trained FCN for the training images were used in selective distillation.
Fig.~\ref{fig:predictions} presents a subregion of a training image which was overlapped with manually annotated character regions or overlapped with predicted character regions at different training epochs.
Although people tried to annotate consistent centers of characters in labeling process, the manual annotations in this sample were lower than actual character centers.
However, because predictions of an initially trained FCN reflect the information of entire GTD, predicted character regions in Fig.~\ref{fig:predictions} were close to character centers compared to the manually annotated positions especially when the training was moderately processed.
After excessive training, the FCN was overfitted to training data, and prediction maps became close to the initial GTD.

To prevent excessive training and overfitting problem in the process of selective distillation, the validation set was used to obtain an optimal distillation epoch.
Distillation epoch is the number of training epoch to obtain a set of modified GTD in the process of selective distillation.
Several sets of modified GTD were generated by using FCN models trained with different epochs, and main training was conducted multiple times with the different sets of modified GTD.
The distillation epoch with the maximum validation accuracy was adopted as an optimal distillation epoch.


\Figure[t!](topskip=0pt, botskip=0pt, midskip=0pt)[width=3.3in]{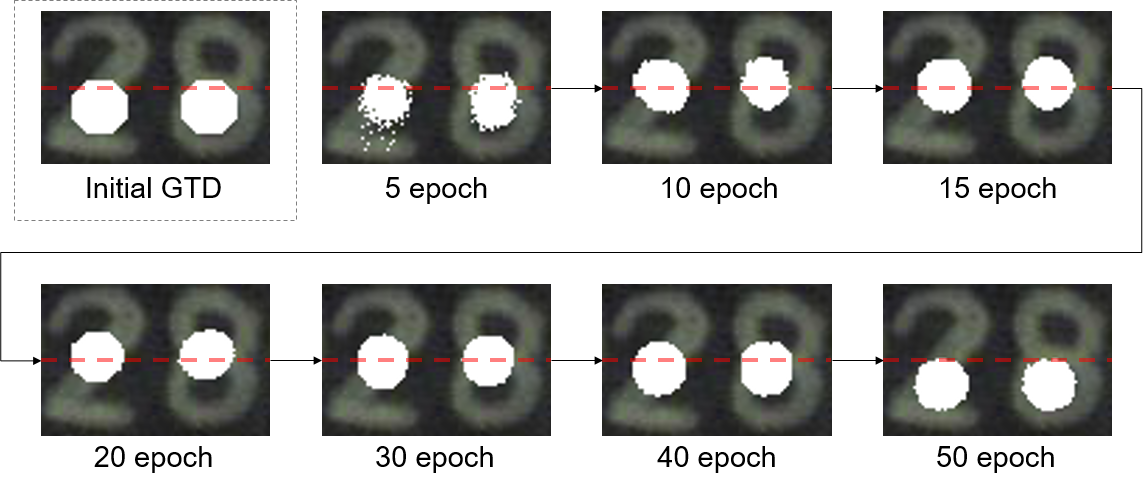}
{A subregion of a training image overlapped with its manually annotated character regions or overlapped with predicted character regions at different training epochs. The vertical centers of the subregions are indicated by red dotted lines.\label{fig:predictions}}

Fig.~\ref{fig:training_procedure} presents the overall training procedure including selective distillation for revising the initial GTD.
In the selective distillation, prediction maps whose SINs are correctly retrieved by the post-processing are selected, and their initial GTD are replaced by the prediction maps.
If there exists an incorrectly recognized SIN in a training image, its prediction map is not selected, and the initial GTD is used in the main training process.
In the main training, the identical FCN model was trained with the modified GTD at the beginning.

\Figure[t!](topskip=0pt, botskip=0pt, midskip=0pt)[width=3.3in]{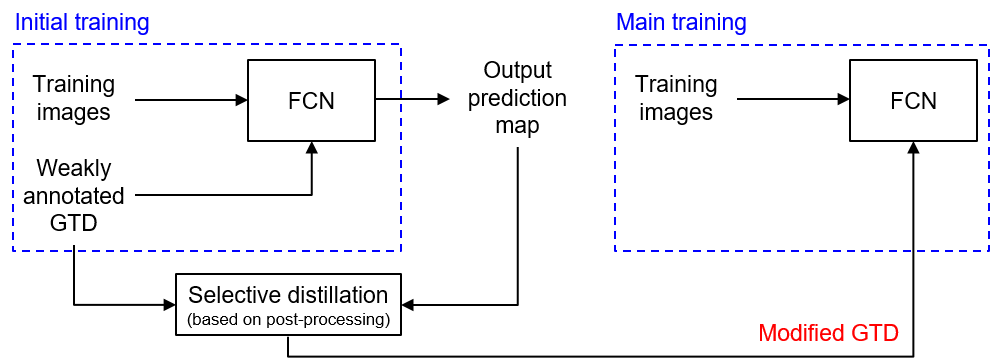}
{Overall training procedure including selective distillation for revising the initial GTD.\label{fig:training_procedure}}

\section{Experiment}

\subsection{Factory Scene Dataset}
Factory scenes were collected as 24-bit color images with the size of $600 \times 960$ at two specific places in an actual steelworks: a slab yard and an entrance of a reheating furnace.
Each factory scene contains a variable number of slabs with handwritten SINs, and 3442 scenes with 4755 slabs were obtained during 181 working days.
The collected data were divided into training, validation and test sets to train and evaluate the proposed algorithm.
Training set and validation set respectively contained 1844 scenes with 2524 slabs and 199 scenes with 285 slabs, and the remaining data were used as a test set to evaluate the algorithm.

\subsection{Experimental environment and evaluation measures}
The proposed algorithm was developed within a hardware environment including Intel core i7-8700K CPU (3.7GHz), 16GB DDR4 RAM and NVIDIA TITAN X (Pascal).
Python and Tensorflow were mainly used for implementation.

In experiments, recognition accuracy was evaluated based on sensitivity, precision and f1-score.
A SIN is composed of nine characters that characterize a slab, and correct recognition of entire characters is required to identify a slab.
The evaluation measures were computed based on the number of correct SINs, and a correct recognition implies that all characters in a SIN are correctly recognized.
In the following experiments, to evaluate recognition accuracy for each case, an optimal training epoch was obtained by investigating the maximum accuracy for the validation set.
The FCN model at the training of the optimal epoch was evaluated for the test set.

\subsection{Initial Training with Weakly Annotated GTD}
The base FCN model presented in Fig.~\ref{fig:FCN_models}(a) was trained for 50 epochs by using the weakly annotated GTD.
One epoch means the training of whole training data once.
In this initial training process, initial learning rate for Adam optimization was set to $10^{-4}$, and the learning rate was decreased by 5\% for each epoch.
Fig.~\ref{fig:initial_training} presents f1-scores for training, validation and test sets during the training of 50 epochs.
The f1-score for the validation set was maximized at the training of 27 epochs, and the corresponding f1-score for the test set was 95.7906\%.
After the training of 30 epochs, test accuracy was not significantly improved, and averaged f1-score for the test set after 30 epochs was 95.4237\%.
This training process took 7.99 h, and the averaged processing time for a test image was 0.1133 s; 0.0772 s and 0.0361 s were respectively taken for the inference of the FCN and post-processing.


\Figure[t!](topskip=0pt, botskip=0pt, midskip=0pt)[width=3.3in]{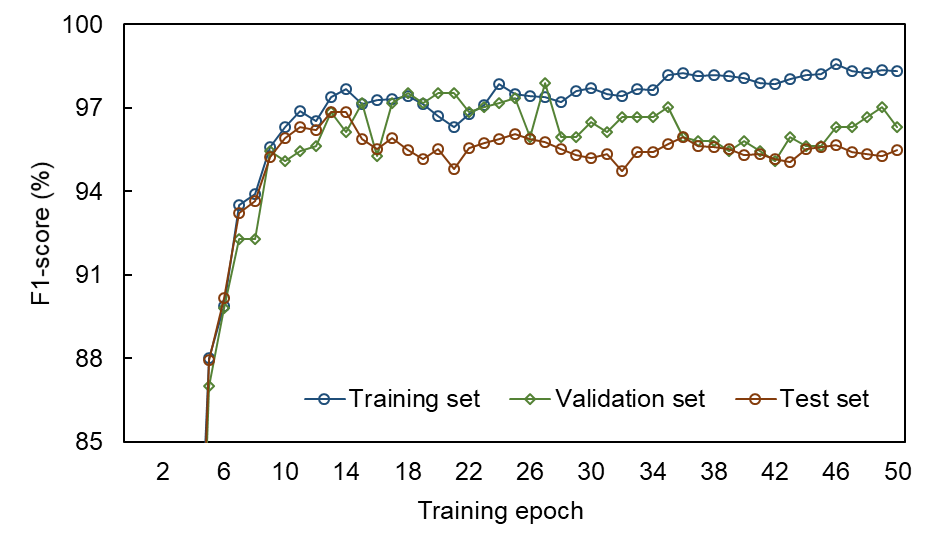}
{Recognition accuracy during initial training with the weakly annotated GTD.\label{fig:initial_training}}

\subsection{Analysis for the FCN models}
\label{sec:analysis_FCNs}

To develop the base FCN model presented in Fig.~\ref{fig:FCN_models}(a), several FCN models were designed and analyzed.
In this section, the performances of \emph{Shallow model} and \emph{Deep model} presented in Fig.~\ref{fig:FCN_models}(b) and Fig.~\ref{fig:FCN_models}(c) are compared to that of the base FCN model to demonstrate appropriate complexity of the base model.
Compared to the base model, \emph{Shallow model} has one less convolution block and deconvolution layer, and \emph{Deep model} has one more convolution block.
A convolution block contains four convolution layers as depicted in Fig.~\ref{fig:FCN_models}.

Table~\ref{tab:FCN_model} summarizes the complexity and accuracy of the three FCN models.
The complexity of a FCN model is closely related to the number of weight layers and the number of trainable parameters, and weight layers indicate convolution and deconvolution layers.
For each model, an optimal training epoch was obtained based on validation accuracy, and sensitivity and precision were measured for the test set at each optimal epoch.
Although \emph{Deep model} has more trainable parameters than the base FCN model, it recorded similar accuracy to the base model due to insufficient training data.
For each case, transfer learning of VGG19 \cite{simonyan2014very}, which was trained for ImageNet dataset \cite{deng2009imagenet}, was used to improve recognition accuracy.
Without using the transfer learning, the base FCN model recorded 95.4265\% and 94.5038\% for sensitivity and precision.
By employing pretrained parameters of VGG19 to initialize parameters in the convolution layers of the base model, sensitivity and precision were improved by 0.4625\% and 1.1885\%.
Although this paper employed VGG19, more recent architectures such U-net \cite{ronneberger2015u} and deeplab \cite{chen2018deeplab} can be used to construct a base network, and selective distillation can be applied to improve the performance in the same way.

\begin{table}
	\centering
	\caption{Numerical Analysis for the Three FCN Models.}
	\resizebox{3.5 in}{!}{
		\begin{tabular}{llll}
			\toprule
			\makebox[3cm][c]{} & \makebox[2cm][c]{Shallow model} & \makebox[2cm][c]{Base model} & \makebox[2cm][c]{Deep model}\\		
			\midrule 
			\makebox[3cm][l]{Number of weight layers} & \makebox[2cm][c]{14} & \makebox[2cm][c]{19} & \makebox[2cm][c]{23}\\	
			\makebox[3cm][l]{Number of parameters} & \makebox[2cm][c]{14,828,876} & \makebox[2cm][c]{26,513,228} & \makebox[2cm][c]{36,542,28}\\	
			\hline \vspace{-6pt}\\
			\makebox[3cm][l]{Training time (h)} & \makebox[2cm][c]{4.75} & \makebox[2cm][c]{7.99} & \makebox[2cm][c]{8.22}\\	
			\makebox[2cm][l]{Test time (s)} & \makebox[2cm][c]{0.0934} & \makebox[2cm][c]{0.1133} & \makebox[2cm][c]{0.1209}\\
			\hline \vspace{-6pt}\\
			\makebox[3cm][l]{Sensitivity (\%)} & \makebox[2cm][c]{93.0627} & \makebox[2cm][c]{\textbf{95.8890}} & \makebox[2cm][c]{95.3751}\\	
			\makebox[3cm][l]{Precision (\%)} & \makebox[2cm][c]{88.7310} & \makebox[2cm][c]{\textbf{95.6923}} & \makebox[2cm][c]{95.1795}\\						
			\bottomrule
		\end{tabular}
	}
	\label{tab:FCN_model}
\end{table}

\subsection{Selective Distillation}
As shown in the Fig.~\ref{fig:predictions}, predictions of an FCN become close to the original GTD as training proceeded.
This section analyzes variations of character regions in modified GTD with respect to different distillation epochs and effectiveness of the selective distillation.

Fig.~\ref{fig:distance_area} presents averaged distance and difference of areas between predicted character regions for correctly recognized SINs and their corresponding GTD.
The averaged distance $D_{avg}$ was computed by
\begin{align}
D_{avg} = \frac{1}{N} \Sigma_{i=1}^N \|y_i - p_i\|^2,
\end{align}
where $y_i$ is the manually annotated center of the \emph{i-th} character, $p_i$ is the centroid of the corresponding predicted character region, and $N$ is the number of correctly recognized characters in the training images.
The averaged difference of areas $A_{avg}$ was computed by
\begin{align}
A_{avg} = \frac{1}{N} \Sigma_{i=1}^N |Y_i - P_i|,
\end{align}
where $A_i$ is the area of the semantic region for the \emph{i-th} character and $P_i$ is the area of the corresponding predicted region.
As shown in Fig.~\ref{fig:distance_area}, $D_{avg}$ and $A_{avg}$ were decreased as training proceeded, and it implies that each prediction map becomes close to its initial ground truth.
Therefore, choosing an appropriate stopping epoch for the initial training is an important issue to prevent an overfitting problem, and we define distillation epoch as the number of training epochs to obtain an initially trained model.


\Figure[t!](topskip=0pt, botskip=0pt, midskip=0pt)[width=3.3in]{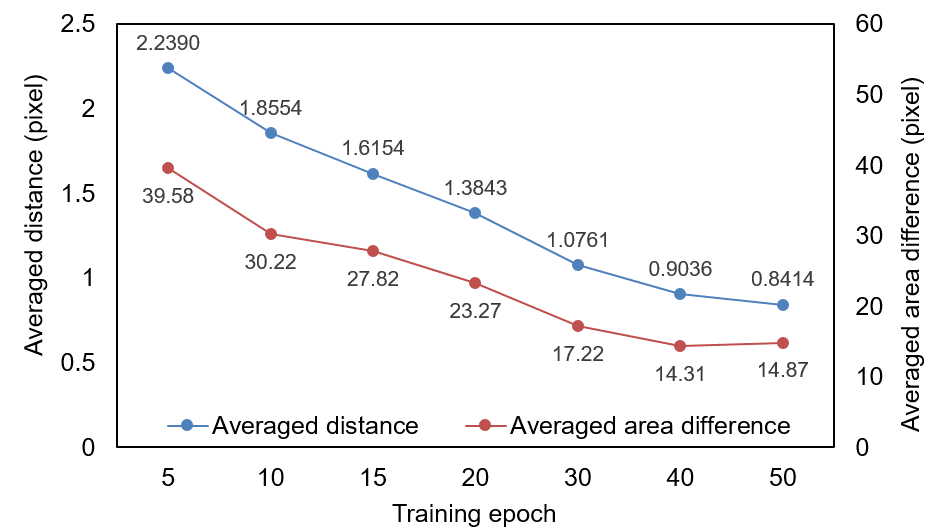}
{Averaged distance and difference of areas between manual annotations and predictions of FCN.\label{fig:distance_area}}

Selective distillation was conducted at various distillation epochs (5, 10, 15, 20, 30, 40, 50 epochs) to generate 7 sets of modified GTD, and main training was conducted for 50 epochs using each set of the modified GTD.
For each FCN model trained using a set of modified GTD, an optimal training epoch was obtained based on validation accuracy, and recognition accuracy was evaluated on the test set at each optimal training epoch.
In Fig.~\ref{fig:selective_distillation}, the maximum validation accuracy and the corresponding test accuracy are presented for each case.
Gray line indicates the test accuracy by using the manually generated GTD, and it was used as the baseline accuracy to compare the performance with selective distillation.
All of the FCN models employing selective distillation outperformed the FCN model trained with the manually generated GTD.

In Fig.~\ref{fig:selection_process}, orange and blue color bars respectively indicate test accuracies with and without the selection process in the selective distillation.
Gray dots indicate distillation ratio for each distillation epoch, where distillation ratio is the ratio of training images whose prediction maps are employed to construct modified GTD.
To obtain accuracies without the selection process, entire predictions were employed as modified GTD.
Fig.~\ref{fig:selection_process} shows that the selection process was effective to improve recognition accuracy for all distillation epochs especially when the initial training was stopped early (5 distillation epoch).

An optimal distillation epoch was determined by using the validation set.
In Fig.~\ref{fig:selective_distillation}, the highest record among the maximum validation accuracies was 98.0736\%, and it was achieved by the FCN model with the distillation epoch of 15.
The corresponding f1-score for the test set was 97.7424\%.
If entire predictions were employed to construct modified GTD by omitting the selection process, f1-score for the test set was decreased to 96.7676\%.
Table~\ref{tab:selective_distillation} summarizes the recognition accuracies by using the three types of GTD: manually generated GTD, modified GTD without selection process and modified GTD with selective distillation.
Fig.~\ref{fig:result_image} presents output prediction maps of the base FCN model trained with the modified GTD whose distillation epoch was 15.
Recognition results for the test scenes are displayed on their output prediction maps.


\Figure[t!](topskip=0pt, botskip=0pt, midskip=0pt)[width=3.3in]{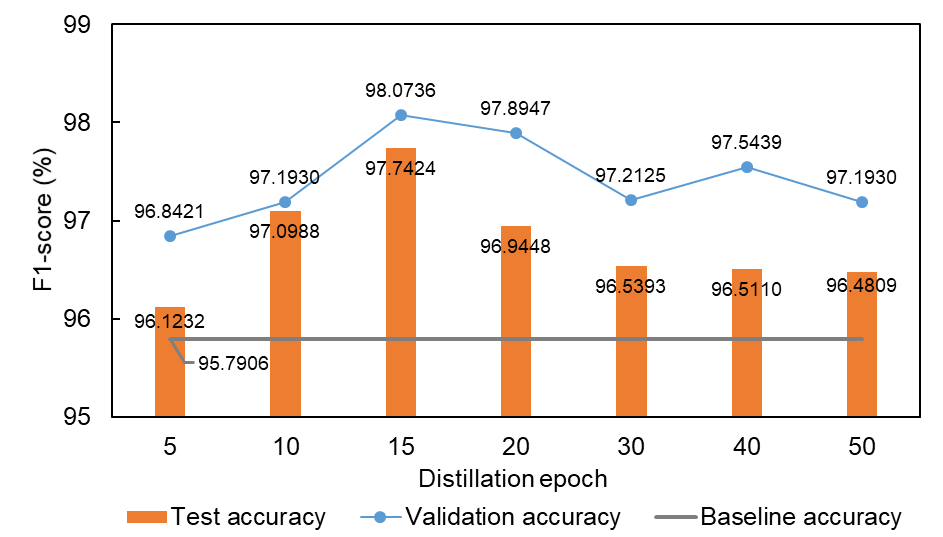}
{Recognition accuracies of FCN models with various distillation epochs.\label{fig:selective_distillation}}

\Figure[t!](topskip=0pt, botskip=0pt, midskip=0pt)[width=3.3in]{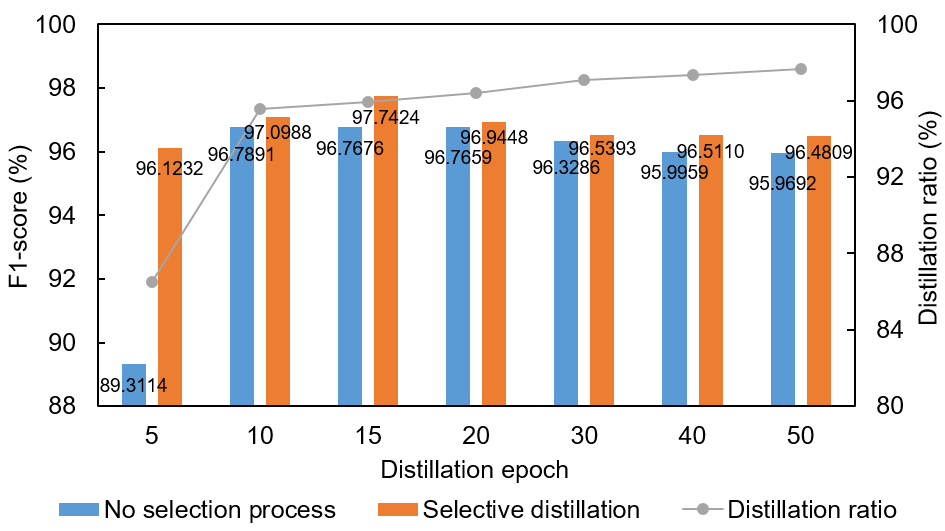}
{Effectiveness of selection process in the selective distillation.\label{fig:selection_process}}


\begin{table*}
	\centering
	\caption{Accuracies of FCNs Trained with Three Types of GTD for Recognizing Handwritten SINs}
	\resizebox{0.9\linewidth}{!}{
		\begin{tabular}{llll}
			\toprule
			\makebox[7cm][l]{GTD type} & \makebox[3cm][c]{Sensitivity (\%)} & \makebox[3cm][c]{Precision (\%)} & \makebox[3cm][c]{F1-score (\%)}\\		
			\midrule
			\makebox[7cm][l]{Manually generated GTD} & \makebox[3cm][c]{95.8890} & \makebox[3cm][c]{95.6923} & \makebox[3cm][c]{95.7906}\\	
			\makebox[7cm][l]{Distilled GTD (without selection)} & \makebox[3cm][c]{96.9168} & \makebox[3cm][c]{96.6189} & \makebox[3cm][c]{96.7676}\\
			\makebox[7cm][l]{Distilled GTD (selective distillation)} & \makebox[3cm][c]{\textbf{97.8931}} & \makebox[3cm][c]{\textbf{97.5922}} & \makebox[3cm][c]{\textbf{97.7424}}\\
			\bottomrule
		\end{tabular}
	}
	\label{tab:selective_distillation}
\end{table*}


\Figure[t!](topskip=0pt, botskip=0pt, midskip=0pt)[width=3.3in]{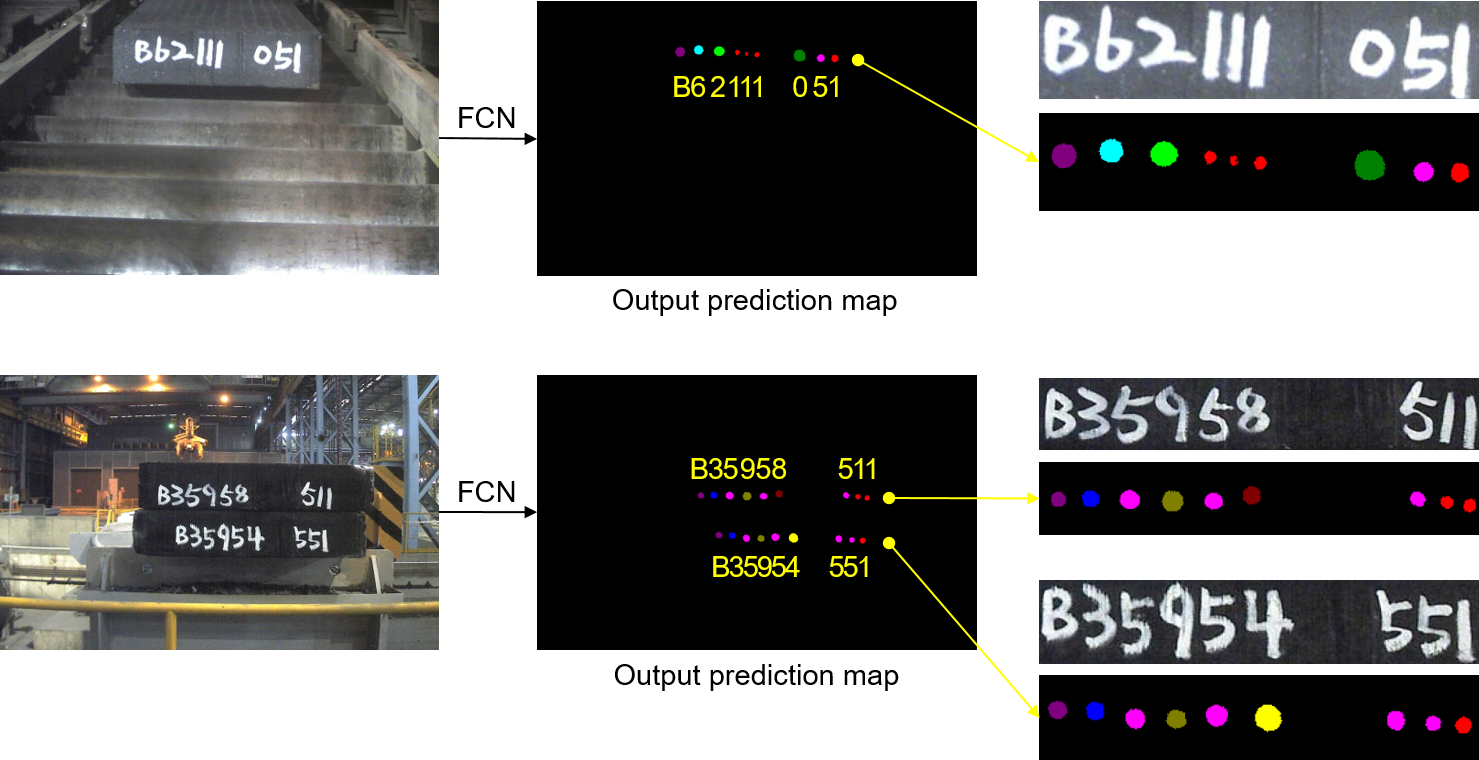}
{Result images for recognizing handwritten SINs in factory scenes. Yellow characters on output prediction maps indicate recognition results.\label{fig:result_image}}

\subsection{Comparison to previous methods}
The previous algorithms for recognizing machine-printed SINs proposed in \cite{lee2017end} and \cite{lee2018recognition} are not applicable to the recognition of handwritten SINs because these methods used features of machine-printed characters such as consistent aspect ratio and similar distances between characters.
However, characters in a handwritten SIN have irregular aspect ratios and varying distances between characters.
Therefore, the proposed algorithm for recognizing handwritten SINs was also applied to the recognition of machine-printed SINs to compare recognition performance with the previous algorithms.

The identical datasets in \cite{lee2017end} and \cite{lee2018recognition} were used for the training and evaluation of the proposed algorithm.
In summary, 1850 scenes with 3749 slabs, 543 scenes with 1102 slabs and 2108 scenes with 4275 slabs were respectively used to construct the training, validation and test sets.
Selective distillation was conducted with the distillation epoch of 10 to revise the weakly annotated GTD.
F1-score for the validation set was maximized at the training of 30 epochs by using the modified GTD, and the corresponding f1-score for the test set was 99.57\%.
The performance of the proposed algorithm and previous methods are summarized in Table~\ref{tab:result_machine}.
Compared to the recent algorithm \cite{lee2018recognition}, error rate of sensitivity was decreased from 0.87\% to 0.43\% in spite of training with weak supervision.
Different to the work by \cite{lee2018recognition} which used region-partitioning method as a post-processing, this paper employed k-means algorithm by representing semantic regions of characters as separated blobs, and it was main reason for the performance improvement.
Fig.~\ref{fig:result_machine} presents result images for recognizing machine-printed SINs.

\Figure[t!](topskip=0pt, botskip=0pt, midskip=0pt)[width=3.3in]{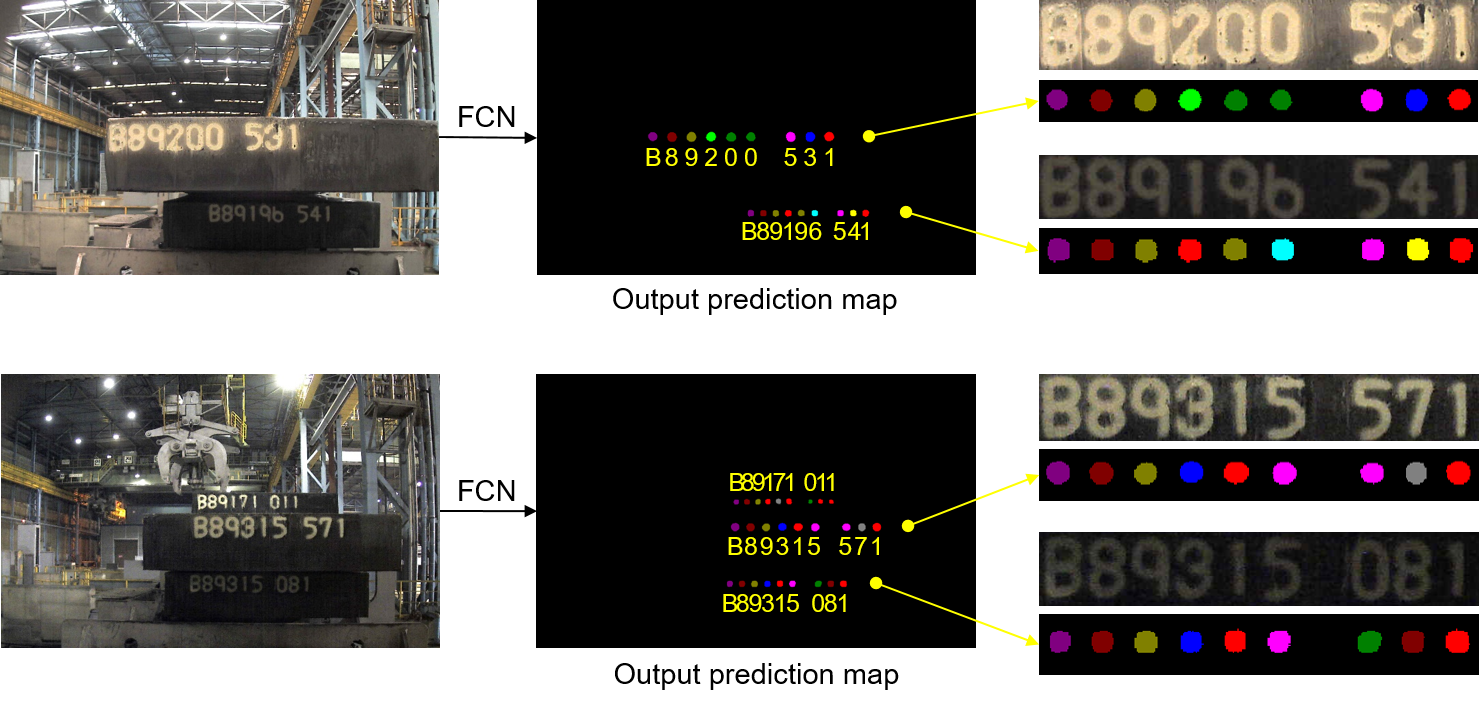}
{Result images for recognizing machine-printed SINs in factory scenes.\label{fig:result_machine}}

\begin{table}
	\centering
	\caption{Performance Comparison with Previous Algorithms}
	\resizebox{1\linewidth}{!}{
		\begin{tabular}{llll}
			\toprule
			\makebox[2.25cm][l]{Reference} & \makebox[2.25cm][c]{Lee \emph{et} al. \cite{lee2017end}} & \makebox[2.25cm][c]{Lee \emph{et} al. \cite{lee2018recognition}} & \makebox[2.25cm][l]{Proposed Method} \\		
			\midrule
			\makebox[2.25cm][l]{Method} & \makebox[2.25cm][c]{CNN} & \makebox[2.25cm][c]{FCN} & \makebox[2.25cm][c]{FCN} \\	
			\midrule			
			\makebox[2.25cm][l]{Sensitivity (\%)} & \makebox[2.25cm][c]{98.36} & \makebox[2.25cm][c]{99.13} & \makebox[2.25cm][c]{\textbf{99.57}} \\
			\makebox[2.25cm][l]{Precision (\%)} & \makebox[2.25cm][c]{99.13} & \makebox[2.25cm][c]{99.16} & \makebox[2.25cm][c]{\textbf{99.57}} \\
			\bottomrule
		\end{tabular}
	}
	\label{tab:result_machine}
\end{table}

\section{Conclusion}
This paper proposes a recognition algorithm for slab identification numbers in factory scenes.
To alleviate manual work in labeling process, weakly annotated data were generated by marking character centroids.
Whereas bounding-boxes, which is a conventional GTD for scene text recognition, require at least a drag-and-drop operation or two clicks to annotate a character location, the weakly annotated GTD requires a single click for recording a character location.
Semantic regions for characters were defined as isolated circular regions to construct 2-dimensional GTD, and these initial GTD were used to train a FCN.
The main contribution of this paper is on the selective distillation for improving the quality of the initial GTD.
Predictions of the initially trained FCN were selectively used to modify the weakly annotated GTD, and the modified GTD were used in the main training procedure.
Experiments were thoroughly conducted on actual industry data to demonstrate the effectiveness of the selective distillation, and the proposed method showed better performance with a large gap compared to the FCN trained with manually annotated GTD.

\EOD

\end{document}